\documentclass{Interspeech}

\interspeechcameraready

\title{Acoustically Precise Hesitation Tagging Is Essential for End-to-End Verbatim Transcription Systems}

\author[affiliation={1}]{Jhen-Ke}{Lin}
\author[affiliation={1}]{Hao-Chien}{Lu}
\author[affiliation={1}]{Chung-Chun}{Wang}
\author[affiliation={1}]{Hong-Yun}{Lin}
\author[affiliation={1}]{Berlin}{Chen}

\affiliation{Speech and Machine Intelligence Lab}{National Taiwan Normal University}{Taiwan}
\email{jacob@ntnu.edu.tw, howchien@ntnu.edu.tw, takala@ntnu.edu.tw, buffett@ntnu.edu.tw, berlin@csie.ntnu.edu.tw}

\usepackage{comment}

\begin{document}

\maketitle

\begin{abstract}
	Verbatim transcription for automatic speaking assessment demands accurate capture of disfluencies, crucial for downstream tasks like error analysis and feedback. However, many ASR systems discard or generalize hesitations, losing important acoustic details. We fine-tune Whisper models on the Speak \& Improve 2025 corpus using low-rank adaptation (LoRA), without recourse to external audio training data. We compare three annotation schemes: removing hesitations (Pure), generic tags (Rich), and acoustically precise fillers inferred by Gemini 2.0 Flash from existing audio-transcript pairs (Extra). Our challenge system achieved 6.47\% WER (Pure) and 5.81\% WER (Extra). Post-challenge experiments reveal that fine-tuning Whisper Large V3 Turbo with the ``Extra'' scheme yielded a 5.5\% WER, an 11.3\% relative improvement over the ``Pure'' scheme (6.2\% WER). This demonstrates that explicit, realistic filled-pause labeling significantly enhances ASR accuracy for verbatim L2 speech transcription.
\end{abstract}

\section{Introduction}

Accurate verbatim transcription for L2 English learners is essential in applications such as automated language-learning feedback, legal proceedings, and medical consultations, where every spoken element, including disfluencies like ``um'' and ``uh'' (filled pauses), must be accurately rendered. However, many automatic speech recognition (ASR) systems discard these tokens during training, since most large-scale online training corpora consist of edited, fluent utterances rather than spontaneous speech. Consequently, original disfluencies are often lost, preventing ASR systems from faithfully transcribing the linguistic content that a speaker actually utters and undermining interpretability.

Speech of L2 learners presents significant challenges to ASR systems due to its wide variability. Lo et al.~\cite{lo2020ntnu} highlighted the diverse speaking characteristics which make L2 speech recognition fall far short of accuracy. Learners frequently exhibit unique pronunciations, non-standard grammar, code-switching, and common disfluencies. Their speech also varies from native norms in acoustics, rhythm, lexical choice, syntax, and pragmatics. Significant intra- and inter-speaker variability, stemming from vocal tract differences and evolving pronunciation, adds further complexity. Moreover, the scarcity of large, publicly available L2 speech datasets with detailed annotations impedes ASR progress. We posit that Whisper's robustness~\cite{radford2022whisper}, derived from massive pre-training, offers a solid foundation that can be enhanced by fine-tuning on dedicated L2 data to address these diverse challenges.

As demonstrated by Bannò et al.~\cite{banno2024towards}, Whisper can be fine-tuned for tasks beyond simple transcription, such as disfluency removal or spoken grammatical error correction, by learning to generate outputs in the desired format. Our work builds upon this observation, focusing on fine-tuning Whisper for \textit{verbatim} transcription that meticulously preserves disfluencies.

The Speak \& Improve (S\&I) Challenge 2025~\cite{qian2024speakimprovechallenge} provided two tracks: a Closed Track, using only official S\&I data and Whisper models, and an Open Track, allowing external resources. The S\&I Corpus 2025~\cite{knill2024speakimprovecorpus}, with around 315-hour audio of L2 English learners (a 55-hour subset of which was manually transcribed with disfluencies), served as a key resource. The official evaluation used word error rate (WER) under a normalized scoring protocol; filled pauses (``um''/``uh'') are removed before scoring, yet we showcase that including them during training substantially benefits lexical content recognition.

On the Closed Track, we selected \textbf{Whisper Large V3} due to its robustness. Pre-trained on approximately 1 million hours of weakly labeled audio and 4 million hours of pseudo-labeled data, it offers superior generalization over Whisper Large V2 (trained on 680,000 hours). Fine-tuning Whisper Large V3 exclusively on S\&I data using low-rank adaptation (LoRA)~\cite{hu2021loralowrankadaptationlarge}, with a targeted post-processing pipeline, reduced the Closed Track baseline WER from 10.84\% to 6.47\%.

On the Open Track, enriching the transcription scheme with precise hesitation tags and punctuation yielded a 5.81\% WER, placing us third overall and demonstrating that acoustically grounded tokens provide more informative supervision.

In post-competition experiments, we focused on balancing accuracy and inference speed using \textbf{Whisper Large V3 Turbo}, a distilled variant. Comparing three transcription schemes—(1) \textbf{Pure} (no hesitations), (2) \textbf{Rich} (generic ``\#'' tags plus punctuation), and (3) \textbf{Extra} (acoustically precise ``um''/``uh'' tokens inferred via Google’s Gemini~\cite{geminiteam2025geminifamilyhighlycapable} 2.0 Flash from S\&I audio-transcript pairs)—we show that explicit, realistic filled-pause labeling yields up to an 11\% relative WER improvement.

Our findings underscore that acoustically precise hesitation tagging is crucial for enhancing recognition accuracy and meeting the authenticity requirements of verbatim ASR.

\section{Methods}

\subsection{Dataset and Preprocessing}

All experiments were conducted on the Speak \& Improve (S\&I) Corpus 2025. This corpus contains 315-hour audio of L2 English learners; a 55-hour subset was manually transcribed and annotated in subtle detail. The annotation process involved multiple phases to capture more nuanced speech characteristics, including disfluencies, errors, and phrase boundaries. Key tags from this process relevant to our work include word-level tags such as \texttt{hesitation} and phrase-level tags such as \texttt{speech-unit-statement} and \texttt{speech-unit-question}.

We generated three transcription variants from these annotations:

\begin{enumerate}[noitemsep, topsep=0pt]
\item \textbf{Pure Transcription}: All extra tags and punctuation removed. This was the target for our Closed Track system.
\item \textbf{Rich Transcription}: This variant maps annotation tags and markers to transcription text according to the rules summarized in Table~\ref{tab:tag-mapping}. This was used in post-challenge experiments.
\item \textbf{Extra Transcription (Acoustically Precise Hesitation Tagging)}: Punctuation included; generic hesitation tags replaced by plausible spoken hesitations (e.g., ``um'', ``uh'') predicted by Google's Gemini 2.0 Flash based on Rich Transcriptions and corresponding audio. This was used for our Open Track system and post-challenge experiments.
\end{enumerate}

\begin{table}[hbt!]
\centering
\caption{Annotation Tag and Mark Mapping for Transcription Variants}
\label{tab:tag-mapping}
\begin{tabular}{p{5cm} p{2cm}}
\toprule
\textbf{Annotation Tag / Mark} & \textbf{Mapped Text} \\
\midrule
\texttt{hesitation}         & \# \\
\texttt{speech-unit-question}    & ? \\
\texttt{speech-unit-statement}  & . \\
\texttt{speech-unit-incomplete} & ... \\
\texttt{partial} (word)         & word- \\
\texttt{backchannel}            & word only (mark dropped) \\
\texttt{disfluency}             & word only (mark dropped) \\
Other detailed annotation tags  & (dropped) \\
\bottomrule
\end{tabular}
\end{table}

\begin{table*}[hbt!]
    \centering
    \caption{Baseline WER of off-the-shelf Whisper models (no fine-tuning) on S\&I Eval Set.}
    \label{tab:baseline}
    \begin{tabular}{lccccc}
        \toprule
        Model Variant  & Parameters & WER (\%)     & Substitutions (\%) & Deletions (\%) & Insertions (\%) \\
        \midrule
        Small          & 244\,M     & 10.7         & 5.0                & 4.4            & 1.3             \\
        Medium         & 769\,M     & 10.4         & 4.4                & 5.1            & \textbf{1.0}    \\
        Large V2       & 1.55\,B    & 9.7          & \textbf{4.2}       & 4.3            & 1.2             \\
        Large V3       & 1.55\,B    & \textbf{9.5} & \textbf{4.2}       & \textbf{4.1}   & 1.2             \\
        Large V3 Turbo & 809\,M     & 9.6          & 4.3                & 4.2            & 1.1             \\
        \bottomrule
    \end{tabular}
\end{table*}

To generate the Extra Transcription data, we utilized Google's Gemini 2.0 Flash model, a multimodal large language model capable of processing both text and audio inputs. The model was tasked with inferring acoustically plausible filled pauses (e.g., ``um'', ``uh'') from the Rich Transcription and the corresponding audio. Specifically, the user prompt provided to Gemini was: \texttt{transcription: RICH TRANSCRIPTION} along with the paired audio file. The system instruction was: \texttt{from the original transcription and speech audio, try to complete the hesitation tags \#, without changing other things.}

For the S\&I Challenge 2025, models were fine-tuned on the combined S\&I training and development sets (~7,100 samples). In the post-challenge experiments, we used only the S\&I training set (~3,900 instances). All evaluations were conducted on the official evaluation set (~3,200 instances).

\subsection{Model Architectures and Fine-Tuning}
Our approach leveraged pre-trained Whisper models, adapting them to the specific task of verbatim L2 speech transcription through fine-tuning. This aligns with the strategy of previous work by Bannò et al.~\cite{banno2024towards}, which demonstrated Whisper's adaptability to various speech processing tasks beyond direct ASR by fine-tuning it on task-specific data. To efficiently fine-tune these large models under resource constraints, rank-stabilized LoRA (rsLoRA)~\cite{kalajdzievski2023rank} was employed. rsLoRA enhances the standard LoRA method by adjusting its scaling mechanism (typically from $\alpha/r$ to $\alpha/\sqrt{r}$, where $r$ is the rank and $\alpha$ is the scaling factor) to improve training stability and performance. In our experiments, it froze all original model weights and injects trainable low-rank matrices into Transformer submodules, drastically reducing the number of tunable parameters. Specific LoRA parameters and training configurations for our experiments are detailed in Section~\ref{sec:finetuning_config_details}.

\subsubsection{Whisper Large V3 (Challenge Tracks)}
\label{subsec:whisper-v3}
For both the Closed and Open Tracks of the challenge, Whisper Large V3 was fine-tuned. This model was selected for its robustness in noisy and accented speech conditions. It was pre-trained on approximately 1 million hours of weakly labeled audio and 4 million hours of pseudo-labeled audio, for a total of 5 million hours. This extensive pre-training offers significant error reduction compared to Whisper Large V2. The specific training configuration for these tracks is detailed in Section~\ref{subsec:challenge_train_cfg}.

For the Closed Track, the ASR model was fine-tuned using the \textbf{Pure Transcription} variant (hesitations and punctuation removed from the target labels).
For the Open Track, the ASR model, initialized from the fine-tuned Closed Track checkpoint, was further fine-tuned using the \textbf{Extra Transcription} variant (with Gemini-enhanced ``um''/``uh'' hesitation tags and punctuation). The LoRA configuration and training hyperparameters for this stage were identical to those described for the Closed Track in Section~\ref{subsec:challenge_train_cfg}.

\subsubsection{Whisper Large V3 Turbo (Post-Challenge)}
In our post-challenge experiments, Whisper Large V3 Turbo was adopted, which is a distilled variant of Whisper Large V3 with its decoder layers reduced from 32 to 4, resulting in 809 million parameters (compared to 1.55 billion in Whisper Large V3). This model achieves faster inference with minimal loss in accuracy. On an NVIDIA A40 GPU, Whisper Small required 65 minutes to transcribe the full S\&I eval set, while Whisper Large V3 Turbo completed the task in 47 minutes. The training configuration for these experiments is detailed in Section~\ref{subsec:post_challenge_train_cfg}.

\subsection{Inference and Post-Processing}
Model outputs were generated in CTM (time-aligned) format and then processed through a normalization pipeline to meet the scoring requirements:
\begin{enumerate}[noitemsep, topsep=0pt]
    \item \textbf{Artifact Filtering}: Tokens with a duration of less than 20~ms and a confidence score below 0.5 were discarded to remove end-of-utterance hallucinations.
    \item \textbf{S\&I Baseline Pipeline Normalizations}: Digits were normalized to words; punctuation and hesitations were removed (as per scoring rules); and other standard preprocessing steps from the S\&I baseline pipeline were applied.
\end{enumerate}
Final transcripts were scored using standard word error rate (WER) with the S\&I evaluation toolkit.

\section{Experiments and Results}

We compare the performance levels among several off-the-shelf Whisper models, our challenge fine-tuning results, and post-challenge experiments under three transcription schemes. All reported WER results reflect the application of our full post-processing pipeline, including artifact filtering.

\subsection{Fine-Tuning Configuration Details}
\label{sec:finetuning_config_details}

\subsubsection{Low-Rank Adaptation (LoRA) Parameters}
\label{subsec:lora_details_cfg}
For all fine-tuning experiments, rank-stabilized LoRA (rsLoRA) was employed. The rsLoRA rank was set to 32, with a scaling factor ($\alpha$) of 8, and a dropout rate of 0.05. rsLoRA was applied to the query, key, value, and output projections of the attention modules, as well as the first and second feedforward layers in the Transformer blocks.

\subsubsection{Challenge Tracks Training Configuration}
\label{subsec:challenge_train_cfg}
Fine-tuning of Whisper Large V3 for the Challenge Tracks (as described in Section~\ref{subsec:whisper-v3}) was performed on a single A40 GPU, using the rsLoRA parameters specified in Section~\ref{subsec:lora_details_cfg}. A per-device batch size of 32 was used with gradient accumulation over 32 steps, resulting in an effective batch size of 1,024 samples per step. The learning rate was $7 \times 10^{-5}$, with AdamW~\cite{loshchilov2018decoupled} optimizer parameters of $\beta_1 = 0.9, \beta_2 = 0.98$, and $\epsilon = 1 \times 10^{-6}$. A weight decay of 0.01 was applied. The model was trained for 28 steps (each comprising 1,024 samples), corresponding to approximately 4.5 epochs on the combined S\&I training and development data.

\subsubsection{Post-Challenge Experiments Training Configuration}
\label{subsec:post_challenge_train_cfg}
In all post-challenge experiments, fine-tuning of Whisper Large V3 Turbo was conducted on a single NVIDIA A40 GPU using the rsLoRA parameters specified in Section~\ref{subsec:lora_details_cfg}. The AdamW optimizer was employed, with hyperparameters set to $\beta_1 = 0.9, \beta_2 = 0.98$, and $\epsilon = 1 \times 10^{-6}$, along with a weight decay of 0.01. The learning rate was fixed at $7 \times 10^{-5}$. Each training update was based on a batch of 32 samples, with no gradient accumulation (i.e., gradient accumulation steps were set to 1). Training proceeded for a total of 732 steps, which corresponds to approximately 6 epochs over the S\&I training set only. Model evaluation and checkpointing were performed every 122 steps to facilitate monitoring and model selection.

\subsection{Baseline Model Evaluation}
Table~\ref{tab:baseline} reports the Word Error Rate (WER) of un-fine-tuned Whisper variants on the S\&I 2025 evaluation set. Note that our measured WER for Whisper Small (10.7\%) is slightly lower than the official Closed Track baseline (10.84\%); the 0.14 percentage point (pp) difference is attributed to the artifact filtering step in our post-processing.

\subsection{Challenge Fine-Tuning}
For the Challenge Tracks, Whisper Large V3 was fine-tuned using LoRA on the combined official S\&I training/development splits, with the training configuration detailed in Section~\ref{subsec:challenge_train_cfg}. Despite a large effective batch size (1,024) and only 28 updates, the following WERs were achieved:
\begin{itemize}[noitemsep, topsep=0pt]
    \item Closed Track WER (Pure Transcription) = 6.47\%
    \item Open Track WER (Extra Transcription) = 5.81\%
\end{itemize}
Our rankings are shown in Table~\ref{tab:closed-rankings} and Table~\ref{tab:open-rankings}.

\begin{table}[hbt!]
  \centering
  \caption{S\&I Challenge ASR Task Closed Track Rankings}
  \label{tab:closed-rankings}
  \begin{tabular}{lcc}
    \toprule
    Rank & Team & WER (\%) \\
    \midrule
    \bfseries 1 & \bfseries E (ours) & \bfseries 6.47 \\
    2 & D & 8.18 \\
    3 & F & 10.83 \\
    4 & G & 12.09 \\
    5 & H & 19.14 \\
    \bottomrule
  \end{tabular}
\end{table}

\begin{table}[hbt!]
  \centering
  \caption{S\&I Challenge ASR Task Open Track Rankings}
  \label{tab:open-rankings}
  \begin{tabular}{lcc}
    \toprule
    Rank & Team & WER (\%) \\
    \midrule
    1 & D & 5.05 \\
    2 & F & 5.05 \\
    \bfseries 3 & \bfseries E (ours) & \bfseries 5.81 \\
    4 & G & 8.20 \\
    5 & H & 10.66 \\
    \bottomrule
  \end{tabular}
\end{table}

\begin{table*}[hbt!]
    \centering
    \caption{Post-challenge WER for Whisper Large V3 Turbo under different transcription schemes on S\&I Eval Set.}
    \label{tab:post}
    \begin{tabular}{lccccc}
        \toprule
        Transcription Scheme & WER (\%)     & $\Delta$ vs.\ Pure (\%) & Substitutions (\%) & Deletions (\%) & Insertions (\%) \\
        \midrule
        Pure                 & 6.2          & -                       & 3.5                & 1.6            & 1.1             \\
        Rich                 & 7.2          & +16.1                   & 3.7                & 2.5            & \textbf{1.0}    \\
        Extra                & \textbf{5.5} & \textbf{-11.3}          & \textbf{3.4}       & \textbf{1.2}   & \textbf{1.0}    \\
        \bottomrule
    \end{tabular}
\end{table*}

\subsection{Post-Challenge Experiments}
To ensure better model convergence, Whisper Large V3 Turbo was subsequently fine-tuned with a more balanced update schedule on the S\&I training set only, as detailed in Section~\ref{subsec:post_challenge_train_cfg}. Three transcription schemes were compared: \textbf{Pure}, \textbf{Rich}, and \textbf{Extra}.
Table~\ref{tab:post} summarizes the WER under each scheme. The results demonstrate that acoustically precise hesitation tagging (``Extra'') consistently outperforms both omission (``Pure'') and generic tagging (``Rich''). The ``Extra'' scheme reduced the WER to 5.5\%, an improvement of 11.3\% relative to the ``Pure'' scheme and 23.6\% relative to the ``Rich'' scheme.

\section{Discussion}

Our post-challenge experiments demonstrate that the ``Extra'' transcription scheme—where acoustically precise filled pauses (e.g., ``um''/``uh'') are labeled—yields the lowest WER (5.5\%) on Whisper Large V3 Turbo, an improvement of 11.3\% relative to the ``Pure'' scheme (Table~\ref{tab:post}). This confirms that preserving real filled-pause tokens, rather than omitting them or using generic tags, strengthens the model's alignment between acoustic patterns and transcript output. This finding resonates with the work of Bannò et al.~\cite{banno2024towards}, who showed that Whisper can be effectively fine-tuned to produce models that incorporate disfluencies (\texttt{Whisper\textsubscript{dsf}}) based on the provided training data.

The challenges of L2 learner speech ASR, as highlighted by Lo et al.~\cite{lo2020ntnu}, particularly concern data scarcity and high acoustic variability. While Lo et al. employed explicit data augmentation techniques like speed perturbation~\cite{ko15_interspeech} and SpecAugment~\cite{park19e_interspeech} to increase the \textit{quantity} of training data, our work leverages the inherent robustness of the large pre-trained Whisper model and focuses on enhancing the \textit{quality and specificity of the fine-tuning data}—specifically, through the detailed and acoustically precise annotation of disfluencies using Gemini. The success of our ``Extra'' scheme underscores the importance of such high-fidelity annotations for achieving robust verbatim transcription, especially for challenging ASR tasks involving nuanced speech phenomena like hesitations.

We employed Google's Gemini 2.0 Flash purely as an offline labeling tool because manual annotation of fine-grained disfluencies across thousands of utterances is prohibitively expensive in terms of human labor and time. This approach to data enrichment is a practical solution to the data bottleneck often faced in specialized ASR tasks. Compared to the high cost of human labelers, we spent \$5 USD to label all the training and development set data. While Bannò et al.~\cite{banno2024towards} noted that their end-to-end system was limited by the availability of task-specific annotated speech data, our use of an external multimodal large language model (MLLM) for hesitation annotation offers a scalable method to generate the necessary supervision for verbatim transcription.

The improvement seen with the ``Extra'' scheme suggests that providing the model with more explicit and acoustically grounded targets for filled pauses allows it to learn more nuanced representations of these events, leading to better overall recognition. This is more effective than using generic ``\#'' tags, which might be too abstract or conflate different types of non-lexical vocalizations.

Future work will focus on two key areas. First, we plan to apply our LLM-assisted enhancement method to scale up the creation of ``Extra''-style transcriptions by processing larger, existing speech corpora that may currently lack detailed hesitation tags. This will increase the volume of training data with acoustically precise disfluency information. Second, we will explore the potential of smaller, potentially open-source MLLMs, such as Microsoft's Phi-4-multimodal-instruct~\cite{abouelenin2025phi}, for the task of inferring these precise hesitations from audio and text. This could offer a more accessible and privacy-respecting alternative to large proprietary models for data enrichment.

\section{Conclusion}

We have shown that acoustically precise hesitation tagging is a critical component for end-to-end verbatim transcription systems for L2 English learner speech. By fine-tuning Whisper Large V3 Turbo with LoRA and comparing three transcription variants, we find that explicit ``um''/``uh'' labeling using Gemini 2.0 Flash reduces WER to 5.5\%, outperforming both omission and generic tagging approaches. This work builds on the understanding that foundation ASR models like Whisper can be adapted for specialized speech processing tasks and addresses the challenges of L2 learner speech variability through careful, targeted data enrichment. Our findings underscore the value of combining robust acoustic modeling with realistic disfluency annotations for high-fidelity ASR and lay the groundwork for more efficient annotation strategies in future verbatim transcription research.

\bibliographystyle{IEEEtran}
\bibliography{mybib}

\begin{thebibliography}{10}
\providecommand{\url}[1]{#1}
\csname url@samestyle\endcsname
\providecommand{\newblock}{\relax}
\providecommand{\bibinfo}[2]{#2}
\providecommand{\BIBentrySTDinterwordspacing}{\spaceskip=0pt\relax}
\providecommand{\BIBentryALTinterwordstretchfactor}{4}
\providecommand{\BIBentryALTinterwordspacing}{\spaceskip=\fontdimen2\font plus
\BIBentryALTinterwordstretchfactor\fontdimen3\font minus \fontdimen4\font\relax}
\providecommand{\BIBforeignlanguage}[2]{{%
\expandafter\ifx\csname l@#1\endcsname\relax
\typeout{** WARNING: IEEEtran.bst: No hyphenation pattern has been}%
\typeout{** loaded for the language `#1'. Using the pattern for}%
\typeout{** the default language instead.}%
\else
\language=\csname l@#1\endcsname
\fi
#2}}
\providecommand{\BIBdecl}{\relax}
\BIBdecl

\bibitem{lo2020ntnu}
T.-H. Lo, F.-A. Chao, S.-Y. Weng, and B.~Chen, ``The ntnu system at the interspeech 2020 non-native children’s speech asr challenge,'' in \emph{Interspeech 2020}, 2020, pp. 250--254.

\bibitem{radford2022whisper}
A.~Radford, J.~W. Kim, T.~Xu, G.~Brockman, C.~McLeavey, and I.~Sutskever, ``Robust speech recognition via large-scale weak supervision,'' in \emph{International conference on machine learning}.\hskip 1em plus 0.5em minus 0.4em\relax PMLR, 2023, pp. 28\,492--28\,518.

\bibitem{banno2024towards}
S.~Bann{\`o}, R.~Ma, M.~Qian, K.~M. Knill, and M.~J. Gales, ``Towards end-to-end spoken grammatical error correction,'' in \emph{ICASSP 2024-2024 IEEE International Conference on Acoustics, Speech and Signal Processing (ICASSP)}.\hskip 1em plus 0.5em minus 0.4em\relax IEEE, 2024, pp. 10\,791--10\,795.

\bibitem{qian2024speakimprovechallenge}
M.~Qian, K.~Knill, S.~Banno, S.~Tang, P.~Karanasou, M.~J. Gales, and D.~Nicholls, ``Speak \& improve challenge 2025: Tasks and baseline systems,'' \emph{arXiv preprint arXiv:2412.11985}, 2024.

\bibitem{knill2024speakimprovecorpus}
K.~Knill, D.~Nicholls, M.~J. Gales, M.~Qian, and P.~Stroinski, ``Speak \& improve corpus 2025: an l2 english speech corpus for language assessment and feedback,'' \emph{arXiv preprint arXiv:2412.11986}, 2024.

\bibitem{hu2021loralowrankadaptationlarge}
\BIBentryALTinterwordspacing
E.~J. Hu, yelong shen, P.~Wallis, Z.~Allen-Zhu, Y.~Li, S.~Wang, L.~Wang, and W.~Chen, ``Lo{RA}: Low-rank adaptation of large language models,'' in \emph{International Conference on Learning Representations}, 2022. [Online]. Available: \url{https://openreview.net/forum?id=nZeVKeeFYf9}
\BIBentrySTDinterwordspacing

\bibitem{geminiteam2025geminifamilyhighlycapable}
G.~Team, R.~Anil, S.~Borgeaud, J.-B. Alayrac, J.~Yu, R.~Soricut, J.~Schalkwyk, A.~M. Dai, A.~Hauth, K.~Millican \emph{et~al.}, ``Gemini: a family of highly capable multimodal models,'' \emph{arXiv preprint arXiv:2312.11805}, 2023.

\bibitem{kalajdzievski2023rank}
D.~Kalajdzievski, ``A rank stabilization scaling factor for fine-tuning with lora,'' \emph{arXiv preprint arXiv:2312.03732}, 2023.

\bibitem{loshchilov2018decoupled}
\BIBentryALTinterwordspacing
I.~Loshchilov and F.~Hutter, ``Decoupled weight decay regularization,'' in \emph{International Conference on Learning Representations}, 2019. [Online]. Available: \url{https://openreview.net/forum?id=Bkg6RiCqY7}
\BIBentrySTDinterwordspacing

\bibitem{ko15_interspeech}
T.~Ko, V.~Peddinti, D.~Povey, and S.~Khudanpur, ``Audio augmentation for speech recognition,'' in \emph{Interspeech 2015}, 2015, pp. 3586--3589.

\bibitem{park19e_interspeech}
D.~S. Park, W.~Chan, Y.~Zhang, C.-C. Chiu, B.~Zoph, E.~D. Cubuk, and Q.~V. Le, ``Specaugment: A simple data augmentation method for automatic speech recognition,'' in \emph{Interspeech 2019}, 2019, pp. 2613--2617.

\bibitem{abouelenin2025phi}
A.~Abouelenin, A.~Ashfaq, A.~Atkinson, H.~Awadalla, N.~Bach, J.~Bao, A.~Benhaim, M.~Cai, V.~Chaudhary, C.~Chen \emph{et~al.}, ``Phi-4-mini technical report: Compact yet powerful multimodal language models via mixture-of-loras,'' \emph{arXiv preprint arXiv:2503.01743}, 2025.

\end{thebibliography}

\end{document}